\pdfoutput=1
\documentclass[11pt]{article}

\usepackage []{ACL2023}

\usepackage{times}
\usepackage{latexsym}

\usepackage[T1]{fontenc}
\usepackage[utf8]{inputenc}

\usepackage{microtype}

\usepackage{times,latexsym}
\usepackage{url}
\usepackage[T1]{fontenc}
\usepackage{url}
\usepackage{comment}
\usepackage{amssymb}
\usepackage{pifont}

\usepackage{inconsolata}
\usepackage{graphicx}
\usepackage{subfigure}
\usepackage{tabularx}
\usepackage{multirow}
\usepackage{algorithm,algcompatible,amsmath}
\usepackage{mathtools}
\usepackage{xcolor}
\usepackage{enumitem}
\usepackage{marvosym}
\usepackage{textcomp}
\usepackage{array}
\usepackage{booktabs}

\newcommand{\MMQA}{\emph{MMQA}}

\title{MMHQA-ICL: Multimodal In-context Learning for Hybrid Question Answering over Text, Tables and Images}

\author{
Weihao Liu$^{1,3}$
\quad Fangyu Lei$^{1,2}$
\quad Tongxu Luo$^{1,3}$
\quad Jiahe Lei$^{3}$ \\
\bf{
Shizhu He$^{1,2}$
\quad Jun Zhao$^{1,2}$ 
\quad Kang Liu$^{1,2}$
}\\
$^1$Instute of Automation, CAS \quad $^2$University of Chinese Academy of Sciences \quad \\ $^3$University of Science and Technology Beijing \\
\texttt{liuweihao2022@outlook.com \quad leifangyu2022@ia.ac.cn}
\\ \texttt{\{shizhu.he, kliu\}@nlpr.ia.ac.cn}
}
\begin{document}
\maketitle
\begin{abstract}
In the real world, knowledge often exists in a multimodal and heterogeneous form. Addressing the task of question answering with hybrid data types, including text, tables, and images, is a challenging task~(MMHQA). Recently, with the rise of large language models~(LLM), in-context learning~(ICL) has become the most popular way to solve QA problems. We propose \textbf{MMHQA-ICL} framework for addressing this problems, which includes stronger heterogeneous data retriever and an image caption module. Most importantly, we propose a \emph{Type-specific In-context Learning Strategy} for MMHQA, enabling LLMs to leverage their powerful performance in this task. We are the first to use end-to-end LLM prompting method for this task. Experimental results demonstrate that our framework outperforms all baselines and methods trained on the full dataset, achieving state-of-the-art results under the few-shot setting on the MultimodalQA dataset.
\end{abstract}

\section{Introduction}

% figure of task | dataset description
\begin{figure*}[htb]
    \centering
    \includegraphics[width= \textwidth]{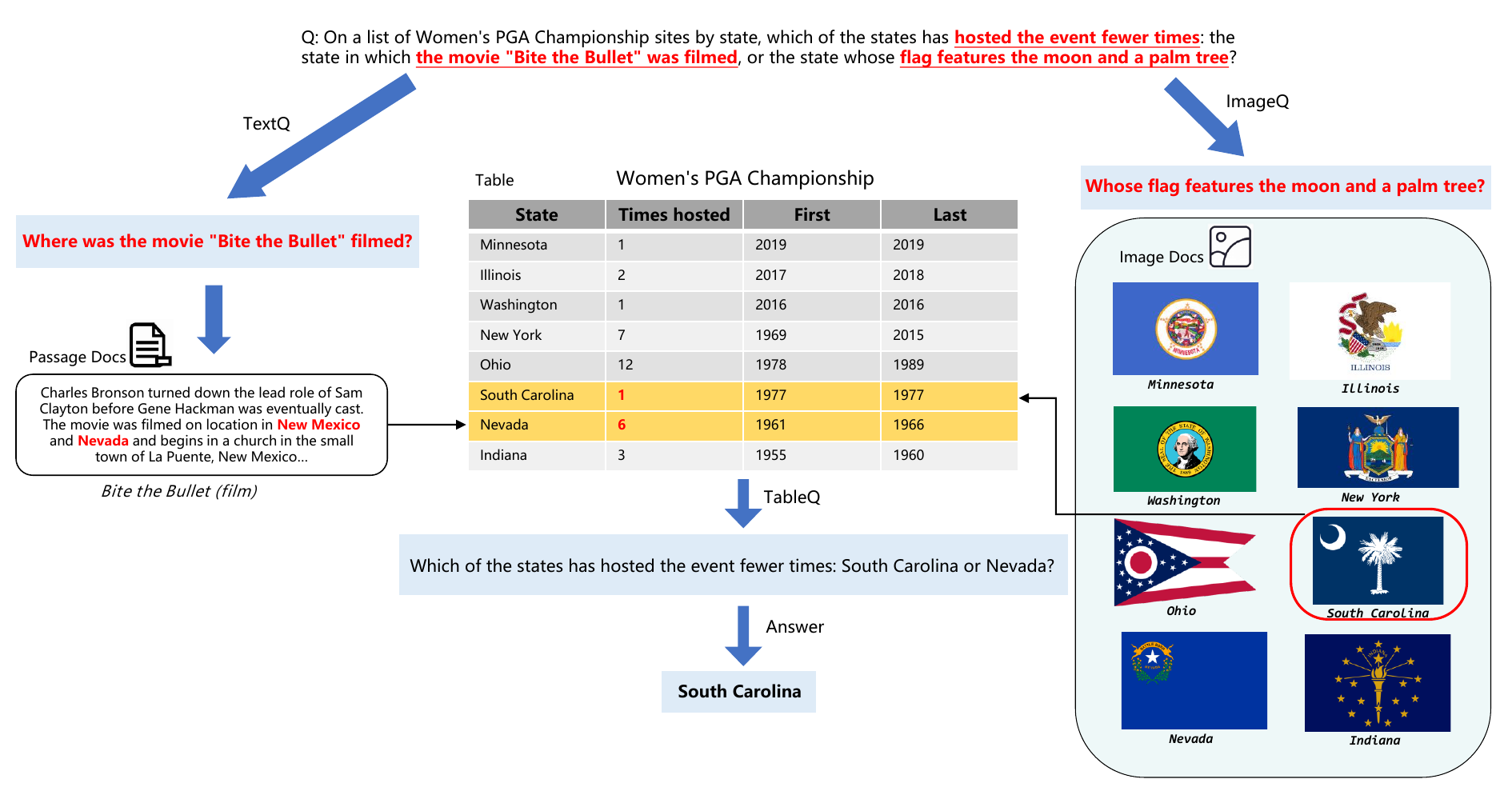}
    \vspace{-2em}
    \caption{An example question of MMQA Dataset}
    \label{fig:example}
    \vspace{-1em}
\end{figure*}

% (A brief introduction to MultimodalQA Tasks, MMQA Dataset, LLM, CoT)
Question answering systems are designed to answer various questions using evidence located in structured knowledge bases (e.g., tables) ~\citep{pasupat2015compositional, yu2018spider} or unstructured texts ~\citep{rajpurkar2016squad}, as well as images~\citep{antol2015vqa}. Considering that many questions need to utilize multiple sources of knowledge jointly in real-world applications, the hybrid form of multimodal hybrid question answering over text, tables and images~(MMHQA) has been proposed and attracted more and more attention~\citep{chen2020hybridqa, zhu2021tat, talmor2020multimodalqa, lei2022answering}.

MMHQA is a challenging task that involves integrating information from various sources with different modals, such as text, tables, and images, to answer questions. This task requires not only the ability to retrieve evidence from multiple modalities but also the capability to use multi-hop reasoning to arrive at the correct answer. As shown in the Figure~\ref{fig:example}, there is a highly complex question. To answer the final question "which of the states has hosted the event fewer times?", we first need to use the image information to answer "Whose flag features the moon and a palm tree?" and the text information to answer "Where was the movie 'Bite the Bullet' filmed?". After we obtain the corresponding intermediate answers, including “South Carolina” and “Nevada”, we then use the information in the table to compare the "hosted times" of these two states and arrive at the final answer, i.e., "South Carolina". This example demonstrates the challenge and complexity of the MMHQA task, which requires models to have the ability to fuse multimodal information and perform multiple rounds of reasoning to complete the task.

%This leads us to two states, "South Carolina" and "Nevada". Then, we use the information 

% 第三段你写一下，说一下先前工作的缺点。在大模型时代之前，Baseline的缺点是采用pipeline方式，会在每一步之间损失大量信息。 MARAG的缺点是什么总结一下。随着大模型的到来，question answering systems have gotten phenomenal improvement，LLMs lack the ability to deal with cross-modal information like images or tables. Binder采用了Text-To-SQL的方式，但是缺点是经常会生成一些错误的SQL，并且标注特定程序example也需要一定的成本？（我理解的不透彻，你自己想一下怎么写）, 

Before the advent of Large Language Models~(LLMs)~\citep{chowdhery2022palm,rae2021scaling,thoppilan2022lamda}, prior works struggled to decompose questions into some sub-questions and solve them step by step, which often leads to the information loss among the steps. While some works~\citep{chen2022murag, yang2022enhancing} attempt to use visual language models to obtain answers directly, these adhoc models require a well-trained multimodal model and extensive fine-tuning on downstream datasets. More importantly, these models lack vital ability to multi-hop reasoning and cannot be easily adopted to other datasets. 

%which often leads to loss of information between the steps. 

% Although some improvement works like ~\citep{chen2022murag, yang2022enhancing} try to use pre-trained multimodal transformer-based encoder and decoder structure to get the answer directly, they need a well pre-trained model and must be fine-tuned on the downstream datasets with lots of data. More importantly, these models lack vital ability to multi-hop reasoning and cannot be easily adopted to other datasets.

With the evolution of large language models~(LLMs) taken place, question answering systems have gotten phenomenal improvements. However, LLMs lack the ability to deal with cross-modal information like images or tables. Motivated by this observation, it has been more and more attractive to study how to excavate the powerful semantic understanding and logical reasoning capabilities of LLMs on cross-modal task. Binder~\citep{cheng2022binding}  develops a new text-to-SQL pipeline that mainly utilizes Codex~\citep{brown2020language} as an SQL generator to transform questions into formal language. However, this approach often fails to answer questions due to SQL syntax errors. Moreover, their method is primarily designed for tables and performs poorly on examples that contain images.

To this end, this paper proposes a \textbf{MMHQA-ICL} framework to solve such MMHQA task. Compared to the fully supervised training approach, our method is more concise. Considering that LLM-generated SQL queries often contain syntax errors~\citep{cheng2022binding}, our framework adopts an end-to-end approach that generates the answer directly without generating intermediate SQL queries. To process images, we propose a LLaVA-based Premium Captioning Module that endows a framework with powerful retrieval and reasoning abilities for image data by generating more semantically abundant image captions. In addition, to relieve the limitation of the number of input tokens and boost the capacity of ICL, we propose a Type-specific In-context Learning Strategy which allows for different prompting strategies to be selected based on the type of question. Experiments demonstrate that our method significantly outperforms previous methods.

% In addition, compared to Binder's use of text-to-SQL methods to generate answers, our end-to-end approach avoids errors caused by SQL syntax. Our proposed Type-specific In-context Learning Strategy allows for different prompting strategies to be selected based on the type of question

In summary, our contributions are as follows:

(1) We are the first to utilize an \textbf{end-to-end} QA method in three-modality hybrid question answering system with \textbf{SOTA} performance.

(2) We propose a \textbf{LLaVA based Premium Captioning Module}, which enhances the quality of image captioning, making it more capable for retrieval and reasoning question.

(3) We introduce a \textbf{Type-specific In-context Learning Strategy} to make LLM more adaptive to different questions, despite the limitation of the amount of input tokens.

\section{Methods}

\subsection{Task Definition}

Given a MultimodalQA task with question and data source in format $T = \left \{\mathcal{Q} ,\mathcal{D} \right \}$, where $\mathcal{D}$ is constructed of $\mathcal{I}, \mathcal{P}$ and $\mathcal{T}$, representing the three-modality data source~(image, passage and table), respectively, MMQA aims to synthesize the information from different data sources and make inferences to give the correct answer $\mathcal{A}$ to the question ultimately.

\subsection{Framework Overview}

\begin{figure*}[htbp]
    \centering
    \includegraphics[width= \textwidth]{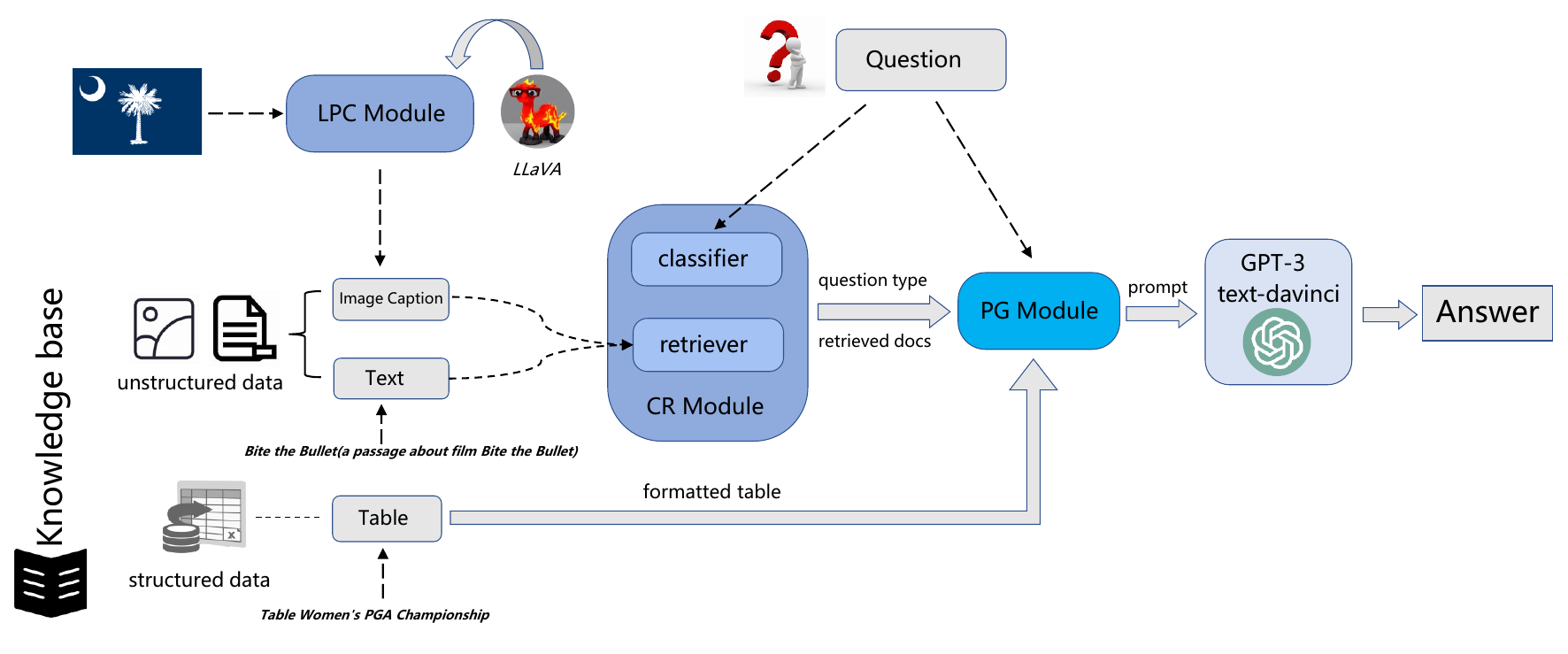}
    \caption{The framework of our method. We first make captions for all images through the LPC~(LLaVA-based Premium Captioning) Module. Then we send questions and related image captions, passages into the CR~(Classifier and Retriever) Module to get \emph{question type} and \emph{retrieved documents}. Next, the PG~(Prompt Generator) Module gathers questions and all retrieved data to build a LLMs input. Finally, LLMs give us the exact answer to the question.}
    \label{fig:framework}
\end{figure*}

As shown in Figure \ref{fig:framework}, we first generate a caption for each image and format the table rows by using tabs so that every data with different modalities can be regarded as texts. Then we put all these questions and texts into a Prompt Generator Module. Subsequently, we gain the input and thus generating final answer of the question through LLM.

% figure of framework

\subsection{LLaVA-based Premium Captioning}
This module is designed to generate textual descriptions of images that can be used as evidence for answering questions in a novel way.

Traditional image captioning models take an image as input to a transformer-based encoder to get the embedding, then send it into a decoder to generate a captioning text. For instance, prior works over MMQA dataset like binder use a vision-text pretrained model OFA ~\citep{wang2022unifying} to generate image captions. However, its evaluation metrics are limited to match-based methods such as BLEU~\citep{papineni2002bleu} and ROUGE-L~\citep{lin2004rouge}, leading into a poor and tedious text which may contain little useful evidence~(e.g., "Plymouth Argyle F.C." $\rightarrow$ "the logo of the fundraiser"). 

Anyway, considering that text-based multimodal question answering systems need a high-quality summary of the exact image, we wish for image captions including more features that are easy to draw people's attention, like colors, characters, costumes, movements, main objects, etc. Consequently, we introduce a LLaVA-based Premium Captioning Module which can generate rich caption texts through LLaVA~\citep{Liu2023VisualIT}. LLaVA is a large Language and Vision Assistant fine-tuned from LLaMA using Visual Instruction Tuning. For example, the picture containing a palm tree in Figure \ref{fig:example} is captioned into text "The image features a blue flag with a white palmetto tree on it, which represents the state of South Carolina."

\subsection{Classifier and Retriever}

In the Classifier and Retriever~(CR) Module, we use DeBERTa-large~\citep{he2023debertav} as the encoder. 

For classifier, unlike the baseline which classifies all questions into 16 possible types, we merely need four types~(image, text, table and compose). We classify all questions through their question text, and finally the accuracy is $96.07\%$ over the validation dataset. 

For retriever, we train two separate models for image and passage. Since images are captioned into texts, we can refer to both images and passages that may be related to a question as documents. Therefore, we can describe the question $\mathcal{Q}$ and related documents $doc$ as several $(\mathcal{Q}, doc)$ pairs. Assuming there are $K_i$ related documents for the $i$-th question $\mathcal{Q}_i$, our goal is to give each $(\mathcal{Q}_i, doc_{ij})$ pair a score $\mathcal{S}_j$ to evaluate their correlation. Such process for $\mathcal{Q}_i$ can be formulated as:

$$\left (\mathcal{S}_1, \mathcal{S}_2, \dots, \mathcal{S}_{K_i} \right) = DeBERTa(V_{i1}, V_{i2}, \dots ,V_{iK_i}) $$
where $V_{ij}$ represents $(\mathcal{Q}_i,doc_{ij})$ and $DeBERTa$ means the DeBERTa-large model. In detail, the $V_{ij}$ is the concatenation of the question, the title and content of the document~(e.g.,[CLS]\textbf{Is it clear or rainy in durban?}[SEP]\textbf{Durban}[SEP]\textbf{The image depicts a lively beach scene with a group of people enjoying their time near the ocean.}[SEP]). 

During training, if there are $n$ gold documents, then the label $y_i$ should be a vector of length $K_i$:

$$y_{ij} = \begin{cases} \frac{1}{n}, & if\ doc_{ij}\ is\ gold\ document \\ 0, & else \end{cases}$$

Our loss function $\mathcal{L}_i$ for the $i$-th question based on $CrossEntropy$ loss can be calculated as follows:

$$\mathcal{L}_i = CrossEntropy(y_i, \hat{y_i})$$
where $\hat{y_i}$ indicates the output $\left (\mathcal{S}_1, \mathcal{S}_2, \dots, \mathcal{S}_{K_i} \right)$ of $DeBERTa$ model .

During evaluation, we take the top-3 of $\mathcal{S}$ documents and calculate the recall to get a better retriever model.

\subsection{Prompt Generator Module}

\paragraph{Type-specific ICL Strategy.} 
% (This may be put at introduction part partially)
% Recently, with the rapid development of large language models, it has been discovered that large language models have powerful emergent capabilities, especially when in-context learning techniques are exerting. At the same time, this technology has also been widely used in common question answering systems. Recently, a new widely respected method called Chain-Of-Thought~\citep{Wei2022ChainOT} has achieved SOTA results on multiple question answering datasets. 

Our work leveraged Chain-of-Thought~(CoT) ~\citep{Wei2022ChainOT} for all questions at the beginning. However, in subsequent experiments, we found that this technology is not universally applicable. Specifically, CoT has a good effect in single-modal tabular questions~(TableQ) and multi-modal compose question answering ~(Compose). But in the task of the single-modal image~(ImageQ) or text~(TextQ) question answering, this technology does not show a strong effect, and even worse. 

After analyzing, we speculated that this is due to the greater emphasis on information recognition and extraction capabilities in single-modal text and image questioning, without the need to generate an explicit chain of reasoning. Meanwhile, due to the limitation of the number of tokens, a single example will consume more tokens when using CoT, resulting in a sharp decrease in the number of examples available for in-context learning, which cannot stimulate the ability of LLM to deal with such problems. Furthermore, the use of CoT would increase inference time exponentially, leading to a low system efficiency. 

Thus, we concluded that not all types of questions are suitable for using the chain of thought technology and we introduced a type-specific prompt generator, which selects different prompt examples depending on the question type for in-context learning. As an example, in the TableQ and cross-modal questions, we use CoT to construct examples. With such an approach, we achieve state-of-the-art performance on the MMQA dataset, showing that type-specific prompt generation can enable LLM to better solve different types of questions. The prompt generation algorithm can be summarized as follows:

\begin{algorithm}[htb]\small
        \renewcommand{\algorithmicrequire}{\textbf{Input:}}
	\renewcommand{\algorithmicensure}{\textbf{Output:}}
	\caption{Prompt Generator Algorithm.} 
	\label{alg1} 
	\begin{algorithmic}[1]
		\REQUIRE question $\mathcal{Q}$, question type $type$, formatted table $\mathcal{T}$, retrieved passages $\mathcal{P}$, retrieved image captions $\mathcal{I}$, examples file $Demos$
		\ENSURE Type-specific prompt $TP$\\
            \emph{Generate specific prompt for different question types}
            \STATE $ TP_{demo} \gets get\_demo\_by\_type(Demos, type)$ 
            \STATE $ TP_{cot} \gets \mathrm{"Please\ answer\ the\ question\ step\ by\ step."}$
            \STATE $ TP_{nocot} \gets \mathrm{"Answer:"}$
		\IF{$ type =  image $}
            \STATE $TP_s \gets \mathcal{Q} + \mathcal{I} + TP_{nocot}$
		\ELSE
            \IF{$ type = text $}
                \STATE $TP_s \gets \mathcal{Q} + \mathcal{P} + TP_{nocot}$
            \ELSE
                \IF{$ type = table$}
                    \STATE $TP_s \gets \mathcal{Q} + \mathcal{T} + TP_{cot}$
                \ELSE
                    \STATE $TP_s \gets \mathcal{Q} + \mathcal{I} + \mathcal{P} + \mathcal{T} + TP_{cot}$
                \ENDIF
            \ENDIF
        \ENDIF
        \STATE $ TP \gets TP_{demo} + TP_s $
	\end{algorithmic} 
\end{algorithm}

% In the Prompt Generator Module, 

\begin{table}[htb]\small
\centering
\resizebox{0.5 \textwidth}{!}{
\begin{tabular}{ll} 
\hline
{\emph{Measurement}} & {\emph{Value}} \\ \hline 
\# Distinct Questions & 29,918 \\
Train multimodal questions & 34.6\% \\
Dev.+test multimodal questions & 40.1\% \\
Train compositional questions & 58.8\% \\
Dev.+test compositional questions & 62.3\% \\
Average question length~(words) & 18.2 \\
%Long questions (more than 20 words) & 29\% \\
Average \# of answers per question & 1.16 \\
List answers & 7.4\%  \\
%Average \# of answers for partial questions & 1.53 \\
List answers per intermediate question & 18.9\%  \\
Average answer length (words) & 2.1 \\
%Answers with more than 1 word & 52.2\% \\
\# of distinct words in questions & 49,649 \\
\# of distinct words in answers & 20,820 \\ 
\# of distinct context tables & 11,022 \\ 
\hline
\end{tabular}}

\caption{Key statistics of MultimodalQA}
\label{tab:key_statistics}

\end{table}

\section{Experiments}

\subsection{Experiment Setup}
\paragraph{Dataset}
We conduct our experiments over MultimodalQA~\cite{talmor2020multimodalqa} dataset. This dataset contains human-annotated multimodal and multi-hop questions over different modalities including tables, text, and images. Wikipedia tables are used as anchors to connect different modalities. The authors first use the template to generate questions and then ask crowd-workers to filter and paraphrase the generated questions. The key statistics of MMQA dataset is listed on Table ~\ref{tab:key_statistics}. Meanwhile, we show the statistics of three single-modal question types and one compose question type in Table ~\ref{tab:statistic-four-types}. We can see that text-based questions and composite questions are the main part with relatively less ImageQ and TableQ, exhibiting a uniform type distribution. It should be note that the author of the MMQA dataset does not disclose the answers to the test dataset or give evaluation channels, so we can only use the development set for our experiments.

\begin{table}[t]
\centering
\resizebox{0.5\textwidth}{!}{
\begin{tabular}{l|l|c}
\hline
\textbf{Type}                     & \textbf{Q\&A}                                                                                                                                                                                                                     & \textbf{\%} \\ \hline
TextQ       & \textit{What was the territorial capital of the territory opposing Ohio} & 31.0    \\ 
                & \textit{in the Toledo War? Detroit} & \\ \hline

TableQ   & \textit{Does the German state Baden-Wurttemberg or Thuringia have more} & 18.3      \\
            & \textit{  residents? Baden-Württemberg} & \\ \hline

ImageQ   & \textit{What weapon is the statue in Nottingham holding? bow} & 15.0      \\ \hline

Compose & \textit{The film that starred Chris Ellison where a man was holding a} & 35.7 \\ 
            & \textit{  newspaper on the poster, was released what year? 1988} & \\ \hline

\end{tabular}}
\caption{Four important types of questions that need to be distinguished in our approach in \MMQA{} with an example and their relative frequency. Previous three rows mean the single-modal questions and the last indicates the cross-modal questions.
%\jb{About TableQ(copmlex): when talking about single-modality questions this was not mentioned, so a bit weird here.}.
}
\label{tab:statistic-four-types}
\end{table}

\paragraph{Implementation Details}

Before the experiment, we leverage LLaVA-13B to generate all captions for each image. Then we regard the image caption as the image itself.

During the classifying and retrieval phase, we used DeBERTa-large as pre-trained language models. For classifier, we confirmed the type of an exact question depending on the highest score among four types. The accuracy of the classifier on the MMQA development set is 96.1\%. For retriever, with k is set to 3, we picked the top k linked passages and images. It identifies all gold documents in 99\% of the paragraphs and 80\% of the images, which has achieved much higher performance on recall concerning passage~($17.3\% \uparrow$) and image~($40\% \uparrow$) compared with prior works.

Stepping into prompt generation stage, we first extracted some demos by question type from a demos file which has been manually built before. Then concatenate them with the information with regard to current question as the input for LLMs.

During the reasoning phase, we used \emph{text-davinci-003} API with the setting \emph{temperature=0.4} to get the final answer to the question. The detailed settings for each question type about \emph{n-shot} and \emph{n} can be found in Table ~\ref{tab:llm_settings}.

% \usepackage{multirow}
\begin{table}[htb]
\resizebox{0.5 \textwidth}{!}{
\begin{tabular}{l|l|c|c|c}
\hline
                         &       & \multicolumn{1}{l|}{n-shot} & \multicolumn{1}{l|}{n-samples} & \multicolumn{1}{l}{max\_generation\_tokens} \\ \hline
\multirow{2}{*}{Image}   & CoT   & 7                           & 1                              & 600                                         \\ \cline{2-5} 
                         & noCoT & 16                          & 8                              & 100                                         \\ \hline
\multirow{2}{*}{Text}    & CoT   & 8                           & 1                              & 600                                         \\ \cline{2-5} 
                         & noCoT & 10                          & 8                              & 100                                         \\ \hline
\multirow{2}{*}{Table}   & CoT   & 6                           & 1                              & 600                                         \\ \cline{2-5} 
                         & noCoT & 9                           & 8                              & 100                                         \\ \hline
\multirow{2}{*}{Compose} & CoT   & 6                           & 1                              & 800                                         \\ \cline{2-5} 
                         & noCoT & 8                           & 8                              & 100                                         \\ \hline
\end{tabular}}

\caption{Number of shots and LLM API parameters for different settings.}
\label{tab:llm_settings}

\end{table}

\subsection{Baselines}

\paragraph{AutoRouting} AutoRouting ~\citep{talmor2020multimodalqa} is a simple approach for answering questions without using cross-modal reasoning by determining the modality where the answer is anticipated to occur and then running the associated single-modality module. 

\paragraph{ImplicitDecomp} ImplicitDecomp ~\citep{talmor2020multimodalqa} introduces a 2-hop implicit decomposition baseline, which is able to automatically determines which portion of the question is pertinent at the present hop. Answers from the first hop are also provided as input in the second hop so that the model can make use of this data and engage in cross-modal reasoning to get the outcome. The model only uses the first hop to determine the response for all single-modality question types (such as \emph{TextQ} and \emph{TableQ}).

\paragraph{MuRAG}

MuRAG ~\citep{chen2022murag} is built on top of a backbone model, which is pre-trained to encode image-text pairs such that they are suitable for both answer generation and retrieval. It takes a query $q$ of any modality as input and retrieves $Top_K$ nearest neighbours from an extracted non-parametric multimodal memory $M$ of image-text pairs. Then the retrievals are combined with the query $q$ as an augmented input $[m_1 ,...,m_k ,q]$, and then fed to the backbone encoder-decoder for answer generation. 

\paragraph{SKURG}

SKURG~\citep{yang2022enhancing} differs from previous methods in that it tightly relates the retrieval and reasoning stages and adapts to arbitrary retrieval hops by unifying the evidence retrieval and answer generation. It uses shared entities to align the sources from various modalities and structured knowledge to map them into a common semantic space.

\paragraph{PReasM} PReasM-Large ~\citep{yoran2022turning} is a pre-trained and fine-tuned language model based on T5-Large for getting strong reasoning ability using different reasoning skills. PReasM also trained two classification models. One for determining whether a question need an image and the other for identifying all gold paragraphs related to a question. They use \emph{ImplicitDecomp} for question that requires images and PReasM model for the remain questions.

\paragraph{Binder} Binder~\citep{cheng2022binding} generates SQL programs and extends the capability of the programming language to solve commonsense problems. The main paradigm of Binder framework is to convert a question into an SQL query through LLM and then execute the SQL to obtain the final answer. Our approach, on the other hand, is different in that we use an \textbf{end-to-end} prompt method, which obtains the answer directly through prompt LLM.

\subsection{Main Results}

We show our \textbf{main results} in Table ~\ref{tab:main_result}. We can see that MMHQA-ICL achieves the best performance on F1 scores among compared baselines, guaranteeing the feasibility of out method. Considering these frameworks without fine-tuning, we can see that we substantially surpasses Binder by 8.7\% on F1 and 3.8\% on EM. Under oracle settings, we furthermore outperforms Binder by 11.4\% on F1 and 6.9\% on EM. It indicates that our method has strong robustness and can be easily adopted to a new dataset. 

In Table ~\ref{tab:subset_result}, we show performance of our method and prior works for the \textbf{questions with different types}. The main advantage of our method is that we are more capable to solve single-modal questions (2.7\% $\uparrow$ on F1), especially text queries(5.7\% $\uparrow$ on F1) compared with SKURG. Predictably, we have much potential towards multi-modal questions since we get a increase of 15.5\% on EM and 15.2\% on F1 with oracle settings. Here oracle settings mean that we use gold types and gold documents as the output of the CR Module.

Additionally, we test \textbf{the effectiveness of our LPC Module} and report the result under \emph{oracle settings} in Table ~\ref{tab:caption_compare}. We can see a sharp drop if we use traditional captioning method, showing that the diversity of semantic information will largely affect the performance on image-related questions. Our LPC Module makes the useful textual description of an image more abundant and achieves huge boost in comparison with traditional works.
% In summary, our MMHQA-ICL method consistently demonstrates strong performance in answering text questions accurately and providing exact answers. It outperforms other methods, both with and without finetuning, and shows further improvement in oracle settings. These results indicate that the MMHQA-ICL model has a robust ability to comprehend and respond to questions effectively.

In summary, our MMHQA-ICL model consistently demonstrates high performance in answering questions across various modalities. It outperforms previous methods and showcases the ability to effectively utilize text, image, and multi-modal information for question answering tasks. Furthermore, the model's performance is significantly enhanced when perfect information is available. Overall, these results validate the effectiveness of the MMHQA-ICL model and LPC Module in handling multi-modal question answering tasks and its potential for achieving state-of-the-art performance in this domain.

% \usepackage{multirow}
\begin{table}[htb]
\centering
\resizebox{0.48 \textwidth}{!}{
\begin{tabular}{lllcc}
\hline
\multicolumn{1}{c}{\textbf{Method}}                                                                 & \multicolumn{1}{c}{} & \multicolumn{1}{c}{} & \textbf{F1}                               & \textbf{EM}                               \\ \hline
\quad \quad \quad \emph{Finetuned}                                                                                          & \multicolumn{2}{c}{}                        &                                           &                                           \\
Implicit-Decomp~\citep{talmor2020multimodalqa}                               &                      &                      & \multicolumn{1}{l}{55.5}                  & \multicolumn{1}{l}{48.8}                  \\
AutoRouting ~\citep{talmor2020multimodalqa}                                  &                      &                      & 49.5                                      & 42.1                                      \\
SKURG~\citep{yang2022enhancing}                                        &                      &                      & 63.8                                      & \textbf{59.4}                                      \\
PReasM-Large~\citep{yoran2022turning}                              & \multicolumn{1}{c}{} & \multicolumn{1}{c}{} & \textbf{65.5}                                      & 59.0                                      \\ \hline
\multirow{2}{*}{\begin{tabular}[c]{@{}l@{}}\quad \emph{Without Finetuning}\\ Binder~\citep{cheng2022binding} \end{tabular}}                &                      &                      &                                           &                                           \\
                                                                                                    &                      &                      & 57.1                                      & 51.0                                      \\
\textbf{MMHQA-ICL}                                                                                 &                      &                      & \textbf{65.8}                            & \textbf{54.8}                            \\ \hline
\multirow{2}{*}{\begin{tabular}[c]{@{}l@{}}\quad \quad \emph{Oracle Settings} \\ $\mathrm{Binder_{oracle}}$~\citep{cheng2022binding} \end{tabular}} &                      &                      & \multicolumn{1}{l}{}     & \multicolumn{1}{l}{}     \\
                                                                                                    &                      &                      & 64.5                     & 58.1                     \\
$\textbf{MMHQA-ICL}_\mathrm{oracle}$                                                                      &                      &                      & \textbf{75.9}            & \textbf{65.0}   \\ \hline
\end{tabular}}

\caption{MMQA F1/EM on development set.}
\label{tab:main_result}

\end{table}

% MHICL-QA results
\begin{table}[htb]
\centering
\resizebox{0.5 \textwidth}{!}{
\begin{tabular}{ccccccccc}
\hline
\multirow{2}{*}{Model}                   & \multicolumn{2}{c}{Text}        & \multicolumn{2}{c}{Image}      & \multicolumn{2}{c}{Single-modal} & \multicolumn{2}{c}{Multi-modal} \\
                                         & EM             & F1             & EM            & F1             & EM              & F1             & EM             & F1             \\ \hline
Q-only                                   & 15.4           & 18.4           & 11.0          & 15.6           & 14.2            & 17.0           & 16.9           & 19.5           \\
AutoRouting                                & 49.5           & 56.9           & 37.8          & 37.8           & 48.9            & 57.1           & 38.2           & 42.1           \\
ImplicitDecomp                           &  -             & -               & -              &  -              & 51.1            & 58.8           & 46.5           & 51.7           \\
MuRAG                                    & 60.8           & 67.5           & \textbf{58.2}          & \textbf{58.2}           & -                &     -           & -               &  -              \\
SKURG                                    & 67.0           & 73.2           & 54.8 & 54.8  & \textbf{66.3}   & 70.2           & \textbf{51.3}  & \textbf{56.4}  \\
\textbf{MMHQA-ICL}                      & \textbf{67.3} & \textbf{78.9}  & 44.2         & 54.1          & 60.5           & \textbf{72.9}  & 46.2          & 55.5          \\
$\textbf{MMHQA-ICL}_\mathrm{oracle}$ & \textbf{72.3} & \textbf{84.9} & \textbf{59.3} & \textbf{67.7} & \textbf{67.2}  & \textbf{79.4} & \textbf{61.7} & \textbf{70.7} \\ \hline
\end{tabular}}

\caption{Performance of our method and related work on the MMQA dev subset.}
\label{tab:subset_result}
\end{table}

\begin{table}[htb]
\centering
\resizebox{0.5\textwidth}{!}{
% \begin{tabular}{llcccccc}
% \hline
% \multicolumn{1}{c}{\textbf{}} & \multicolumn{1}{c}{} & \multicolumn{2}{c}{\textbf{ImageQ}} & \multicolumn{2}{c}{\textbf{ImageListQ}} & \multicolumn{2}{c}{\textbf{Cross-Modal}} \\
% \multicolumn{1}{c}{}          & \multicolumn{1}{c}{} & EM               & F1               & EM                 & F1                 & EM                  & F1                 \\ \hline
% Caption$_\mathrm{LPC}$                        &                      & \textbf{46.1}    & \textbf{50.4}    & \textbf{80.9}      & \textbf{96.0}      & \textbf{61.7}       & \textbf{70.7}      \\
% Caption$_\mathrm{normal}$           &                      & 32.6             & 36.3             & 79.4               & 94.3               & 58.1                & 67.8               \\ \hline
% \end{tabular}
% \begin{tabular}{lcccccc}
% \hline
% \multicolumn{1}{c}{\textbf{}} & \multicolumn{2}{c}{\textbf{ImageQ}} & \multicolumn{2}{c}{\textbf{ImageListQ}} & \multicolumn{2}{c}{\textbf{Cross-Modal}} \\
% \multicolumn{1}{c}{}          & EM               & F1               & EM                 & F1                 & EM                  & F1                 \\ \hline
% Caption$_\mathrm{LPC}$        & \textbf{46.1}    & \textbf{50.4}    & \textbf{80.9}      & \textbf{96.0}      & \textbf{61.7}       & \textbf{70.7}      \\
% Caption$_\mathrm{normal}$      & 32.6             & 36.3             & 79.4               & 94.3               & 58.1                & 67.8               \\ \hline
% \end{tabular}

\begin{tabular}{lcccc}
\hline
\multicolumn{1}{c}{\textbf{}} & \multicolumn{2}{c}{\textbf{ImageQ}} & \multicolumn{2}{c}{\textbf{Cross-Modal}} \\
\multicolumn{1}{c}{}          & EM               & F1               & EM                  & F1                 \\ \hline
Caption$_\mathrm{LPC}$        & \textbf{46.1}    & \textbf{50.4}    & \textbf{61.7}       & \textbf{70.7}      \\
Caption$_\mathrm{normal}$     & 32.6             & 36.3             & 58.1                & 67.8               \\ \hline
\end{tabular}

}

\caption{Results of image-related question types with using LPC captioning or normal captioning method under oracle settings.}
\label{tab:caption_compare}

\end{table}

\begin{table*}[htb] \small
\centering
\begin{tabular}{llcccccccccc}
\hline
\multicolumn{1}{c}{\textbf{}} & \multicolumn{1}{c}{} & \multicolumn{2}{c}{\textbf{Image}} & \multicolumn{2}{c}{\textbf{Text}} & \multicolumn{2}{c}{\textbf{Table}} & \multicolumn{2}{c}{\textbf{Cross-Modal}} & \multicolumn{2}{c}{\textbf{All}} \\
\multicolumn{1}{c}{}          & \multicolumn{1}{c}{} & EM               & F1              & EM              & F1              & EM               & F1              & EM                  & F1                 & EM              & F1             \\ \hline
Diverse-Prompt + noCoT  &                      & \textbf{44.2}    & \textbf{54.1}   & \textbf{67.3}   & \textbf{78.9}   & \textbf{61.3}    & \textbf{76.3}   & 41.3       & 50.2      & \textbf{52.4}   & \textbf{63.2}  \\
Coherent-Prompt + noCoT       &                      & 31.0             & 41.8            & 61.4            & 73.5            & 58.0             & 72.4            & \textbf{42.1}                & \textbf{51.6}               & 48.6            & 59.7           \\ \hline
Diverse-Prompt + CoT    &                      & \textbf{36.9}    & \textbf{46.8}   & \textbf{64.9}   & \textbf{78.7}   & \textbf{63.7}    & \textbf{80.1}   & 46.2       & 55.2      & \textbf{53.0}   & \textbf{64.7}  \\
Coherent-Prompt + CoT         & \multicolumn{1}{c}{} & 31.0             & 42.7            & 63.9            & 76.3            & 59.4             & 73.8            & \textbf{47.4}                & \textbf{56.6}               & 51.6            & 63.0           \\ \hline
\end{tabular}

\caption{Comparison of the performance between using Diverse-Prompt and Coherent-Prompt with CoT or noCoT. Where Diverse-Prompt means using different demos and prompt formats for each question type, Coherent-Prompt means using the same complex prompt for any question.}
\label{tab:ablation1}

\end{table*}

\begin{table*}[htb] \small
\centering
\begin{tabular}{l|clclcccccc}
\hline
\multicolumn{1}{c|}{}                 & \multicolumn{2}{c}{\textbf{Image}}                                    & \multicolumn{2}{c}{\textbf{Text}}                                     & \multicolumn{2}{c}{\textbf{Table}}                                    & \multicolumn{2}{c}{\textbf{Cross-Modal}} & \multicolumn{2}{c}{\textbf{All}}                                      \\
                                      & EM                                & \multicolumn{1}{c}{F1}            & EM                                & \multicolumn{1}{c}{F1}            & EM                                & F1                                & EM                  & F1                 & EM                                & F1                                \\ \hline
\textbf{Partial CoT(best)}            & \multicolumn{1}{l}{\textbf{44.2}} & \textbf{54.1}                     & \multicolumn{1}{l}{\textbf{67.3}} & \textbf{78.9}                     & \multicolumn{1}{l}{\textbf{63.7}} & \multicolumn{1}{l}{\textbf{80.1}} & 46.2    & 55.2      & \multicolumn{1}{l}{\textbf{54.8}} & \multicolumn{1}{l}{\textbf{65.8}} \\
All CoT                               & \multicolumn{1}{l}{36.9}          & 46.8                              & \multicolumn{1}{l}{64.9}          & 78.7                              & \multicolumn{1}{l}{\textbf{63.7}} & \multicolumn{1}{l}{\textbf{80.1}} & \textbf{46.3}       & \textbf{55.3}      & \multicolumn{1}{l}{53.0}          & \multicolumn{1}{l}{64.7}          \\
No CoT                                & \multicolumn{1}{l}{\textbf{44.2}} & \textbf{54.1}                     & \multicolumn{1}{l}{\textbf{67.3}} & \textbf{78.9}                     & 61.3                               & 76.3                               & 41.3                 & 50.2                & 52.4                               & 63.2                               \\ \hline
\quad \quad \emph{oracle settings} & \multicolumn{1}{l}{}              &                                   & \multicolumn{1}{l}{}              &                                   & \multicolumn{1}{l}{}              & \multicolumn{1}{l}{}              &                     &                    &                                   &                                   \\
\textbf{Partial CoT(best)}            & \textbf{59.3}                     & \multicolumn{1}{c}{\textbf{67.7}} & \textbf{72.3}                     & \multicolumn{1}{c}{\textbf{84.9}} & \textbf{65.0}                     & \textbf{80.4}                     & \textbf{61.7}       & \textbf{70.7}      & \textbf{65.0}                     & \textbf{75.9}                     \\
All CoT                               & 40.2                              & \multicolumn{1}{c}{50.3}          & 70.6                              & \multicolumn{1}{c}{84.7}          & \textbf{65.0}                     & \textbf{80.4}                     & \textbf{61.7}       & \textbf{70.7}      & 61.6                              & 73.2                              \\
No CoT                                & \textbf{59.3}                     & \multicolumn{1}{c}{\textbf{67.7}} & \textbf{72.3}                     & \multicolumn{1}{c}{\textbf{84.9}} & 63.1                              & 78.2                              & 54.6                & 64.2               & 61.8                             & 72.9                              \\ \hline
\end{tabular}
\caption{F1 and EM results on the development set considering using CoT or not. Partial CoT represents only using CoT in questions demanding table or multi-modal information. Respectively, All CoT means using CoT for all questions and No CoT means no CoT strategy entirely.}

\label{tab:ablation2}
\end{table*}

\subsection{Ablation Study}

We conduct ablation experiments to verify \textbf{the effectiveness of our unique Type-specific ICL Strategy}, which are presented in Table ~\ref{tab:ablation1} and Table ~\ref{tab:ablation2}.

\paragraph{Diverse-Prompt or Coherent-Prompt.}
In Table ~\ref{tab:ablation1}, we compare Diverse-Prompt with Coherent-Prompt which using the same complex prompt for all types. Considering the situation of noCoT, we can observe that using Diverse-Prompt boosts 13.2\% on EM and 12.3\% on F1 for Image as well as roughly 4\% increase over all questions. Such enhancement indicates that utilizing different demos for different question types does work effectively. Under CoT settings, the phenomenon is similar. Moreover, it should be clarified that due to wrong classification result that classifies cross-modal question into single-modal, Diverse-Prompt gets a lower scores than Coherent-Prompt in cross-modal questions, but it is not the main issue.

\paragraph{Using CoT or noCoT.} We furthermore investigate the influence of using CoT or not. Table ~\ref{tab:ablation2} demonstrates the results under all six cases. When using CoT in all types, the inference time increases heavily but we can see that performance drops by 7.3\% under normal settings and 19.1\% under oracle settings in Image, 2.4\% and 1.7\% in Text and eventually causing 1.8\% and 3.4\% decline over the whole dataset. Without using any CoT under normal settings, it causes 4.9\% down on EM and 5\% down on F1 in Cross-Modal questions, finally leading to 2.4\% decline over all questions. Such phenomenon is also more obvious under oracle settings.

\subsection{Case Study}

\begin{figure*}[htbp]
    \centering
    \includegraphics[width= \textwidth]{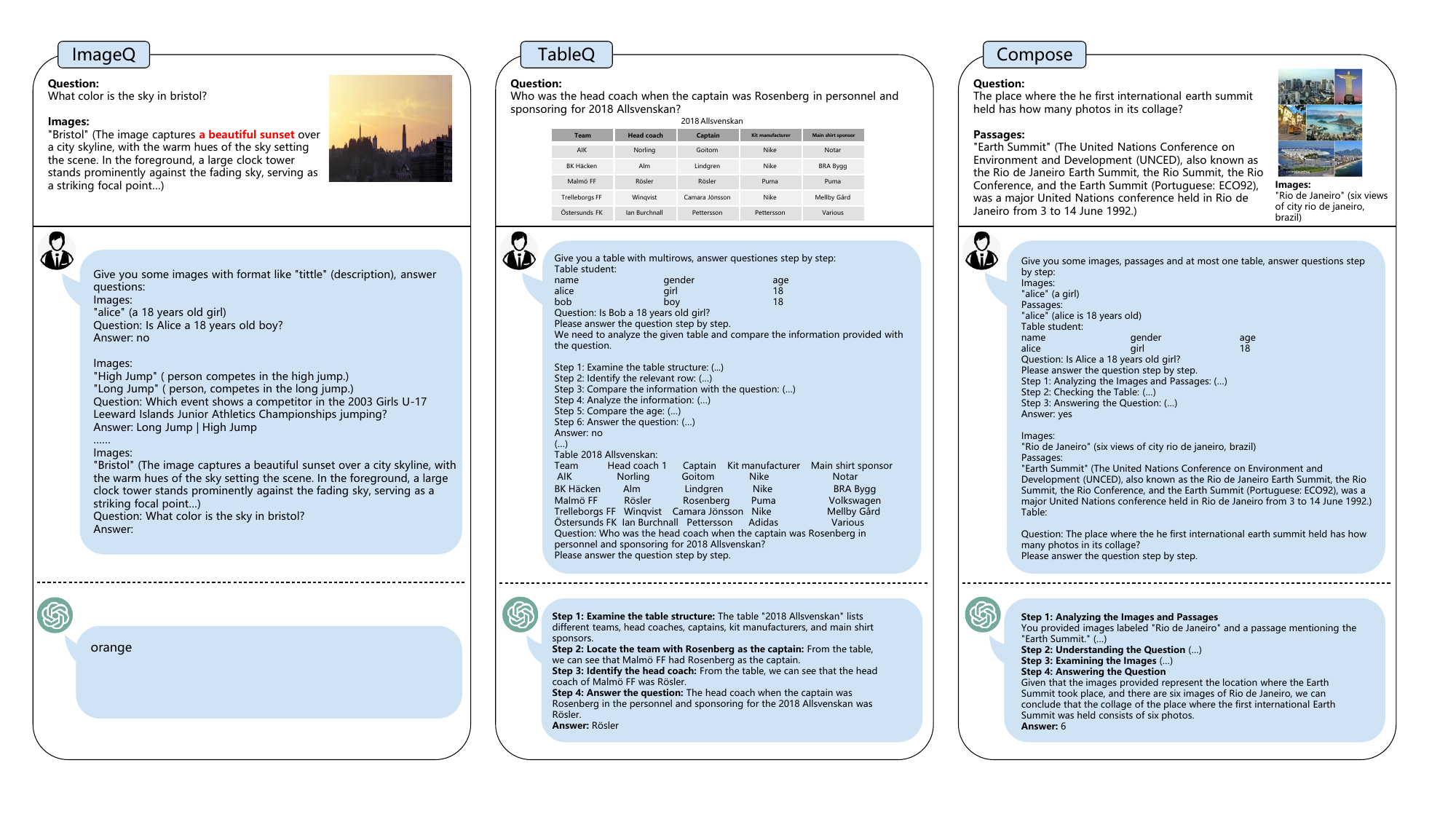}
    \caption{Case study when using MMHQA-ICL}
    \label{fig:case_study}
\end{figure*}

We list three examples from MMQA in Figure ~\ref{fig:case_study}. The left example is a question only related to visual modality. In this case, we only select similar visual questions as demos without CoT and only use the image part of the related documents to build prompt. As we can see, LLM can infer that the sky is in orange while the image caption only provide a sunset background. 

The middle example is a table-based question so we using demos that contains each reasoning steps. Motivated by these demos, LLM derives correct answer following similar steps:

1) Examine the table structure. 

2) Locate the information needed in the table. 

3) Identify the required data. 

4) Answer the question. 

The right example is a compose question containing both passage and image. To solve this question, LLM is supposed to synthesize the information from two modalities. Such process is suitable for using CoT. Thus, LLM analyses the question and related data step by step and responds with the right answer at last.

% Thus LLM analyses the question and related data step by step, respond with a right answer at last.

% figure of case study

\section{Related Work}

\paragraph{Multimodal Hybrid Question Answering} 
Multimodal hybrid question answering has been extensively studied in recent years. For image-text hybrid question answering, MuMuQA~\cite{reddy2021mumuqa}, ManyModelQA~ \cite{hannan2020manymodalqa}, and MIMOQA ~\cite{singh2021mimoqa} have been introduced, which provide questions that require reasoning over images and explicitly provided text snippets~\citep{wang2023hoppg}. Another branch of the task is TextTableQA~\citep{zhu2021tat, chen2020hybridqa, wei2023multi, lei2023s3hqa}, which aims to obtain answers by utilizing hybrid data types of text and tables, including structured database tables~\citep{yu2018spider} and unstructured spreadsheets~\citep{zhao-etal-2022-multihiertt}. Our work is distinct from prior studies as we address three types of data, namely tables, text, and images, and achieve state-of-the-art results on the MultimodalQA dataset. In contrast, previous research mainly focused on two types.

\paragraph{In-Context Learning} 
Large language models such as GPT-3 exhibit impressive few-shot learning ability~\citep{liu2023pre, dong2022survey}, requiring only a few questions and answers as prompts in the context without the need for finetuning on a dataset of training examples. In addition to text-based QA~\citep{Wei2022ChainOT}, in-context learning has also shown promising results in tableQA~\cite{cheng2022binding, chen2022large, ye2023large} and visual~\citep{zhang2023multimodal} QA tasks. However, our proposed MMHQA-ICL framework distinguishes itself from these works by incorporating a type-specific in-context learning strategy, which can effectively handle mixed QA data of three different types: images, text, and tables.

\section{Conclusion}

In this paper, we proposed a MMHQA-ICL framework for the challenging task of hybrid question answering over text, tables, and images. Our framework includes a stronger heterogeneous data retriever and an image captioning module, as well as a type-specific in-context learning strategy that enables LLMs to leverage their powerful performance in this task. We are the first to use an end-to-end prompting method for this task, and our experimental results demonstrate that our framework outperforms all baselines and methods trained on the full dataset, achieving state-of-the-art results under the few-shot setting on the MultimodalQA dataset.

\section*{Limitations}

Since the multimodal hybrid question answering problem has only one dataset MultimodalQA, our model has experimented on only one dataset. This may lead to a lack of generalizability of our model. Transparency and interpretability are important in multi-hop question answering. While our model achieves the best results, the model does not fully predict the reasoning path explicitly. In future work, we will design more interpretable MMHQA models.

\bibliography{anthology,custom}
\bibliographystyle{acl_natbib}

\end{document}